\documentclass[10pt,twocolumn,letterpaper]{article}

\usepackage{cvpr}              
\usepackage{caption}
\usepackage{graphicx}
\usepackage{float} 
\usepackage{subcaption}
\usepackage{paralist}
\usepackage{multirow}
\usepackage{booktabs}
\usepackage{pifont}
\usepackage{bbding}

\makeatletter
\def\hlinew#1{%
	\noalign{\ifnum0=`}\fi\hrule \@height #1 \futurelet
	\reserved@a\@xhline}
\makeatother%

\usepackage[dvipsnames]{xcolor}

\definecolor{cvprblue}{rgb}{0.21,0.49,0.74}
\usepackage[pagebackref,breaklinks,colorlinks,citecolor=cvprblue]{hyperref}

\def\paperID{2} 
\def\confName{CVPRW}
\def\confYear{2024}

\title{UAV-Rain1k: A Benchmark for Raindrop Removal from UAV Aerial Imagery}

\author{Wenhui Chang$^{1}$\quad Hongming Chen$^{1}$\quad Xin He$^{2}$\quad Xiang Chen$^{3}$\thanks{Corresponding author.} \quad Liangduo Shen$^{4}$$^{*}$\\
$^{1}$Shenyang Aerospace University \quad
$^{2}$Naval Aviation University \\
$^{3}$Nanjing University of Science and Technology \quad
$^{4}$Zhejiang Ocean University 
\\
}

\begin{document}
\maketitle
\begin{abstract}
Raindrops adhering to the lens of UAVs can obstruct visibility of the background scene and degrade image quality. Despite recent progress in image deraining methods and datasets, there is a lack of focus on raindrop removal from UAV aerial imagery due to the unique challenges posed by varying angles and rapid movement during drone flight. To fill the gap in this research, we first construct a new benchmark dataset for removing raindrops from UAV images, called UAV-Rain1k. In this paper, we provide a dataset generation pipeline, which includes modeling raindrop shapes using Blender, collecting background images from various UAV angles, random sampling of rain masks and etc. Based on the proposed benchmark, we further present a comprehensive evaluation of existing representative image deraining algorithms, and reveal future research opportunities worth exploring. The proposed dataset is publicly available at \url{https://github.com/cschenxiang/UAV-Rain1k}.
\end{abstract}

\section{Introduction}
\label{sec1}

As unmanned aerial vehicles (UAVs) are increasingly employed for tasks such as surveillance, mapping, and environmental monitoring, the quality of captured imagery becomes paramount. Aerial images are susceptible to degradation in clarity, color distortion, and loss of information due to adverse weather conditions and other factors during the acquisition process~\cite{ye2021deep}. For example, during rainy weather, raindrops adhering to the lens of UAVs can significantly degrade image quality, hindering visibility and affecting the accuracy of downstream image processing tasks. Thus, developing effective raindrop removal techniques tailored to UAV imagery is essential for enhancing the reliability and usability of UAV-based applications~\cite{chen2022unpaired,ye2021closing,wei2021deraincyclegan}.

In recent years, notable progress has been made in image deraining, propelled by the advent of effective image priors and deep learning networks \cite{chen2023towards,chen2020multi}. To better address the problem of removing raindrops, Qian et al. \cite{qian2018attentive} constructed the first raindrop removal dataset, comprising 1,119 pairs of images featuring raindrops against varied backgrounds. These images are captured using a camera equipped with two aligned pieces of glass, one sprayed with water while the other remained clean. Later, Quan et al. \cite{quan2021removing} presented a mixed rain dataset, namely RainDS, which contains synthetic and real-world raindrops created by manually mimicking rainfall with a sprinkler.
\begin{figure}[!t]
\centering
\includegraphics[width=1.0\columnwidth]{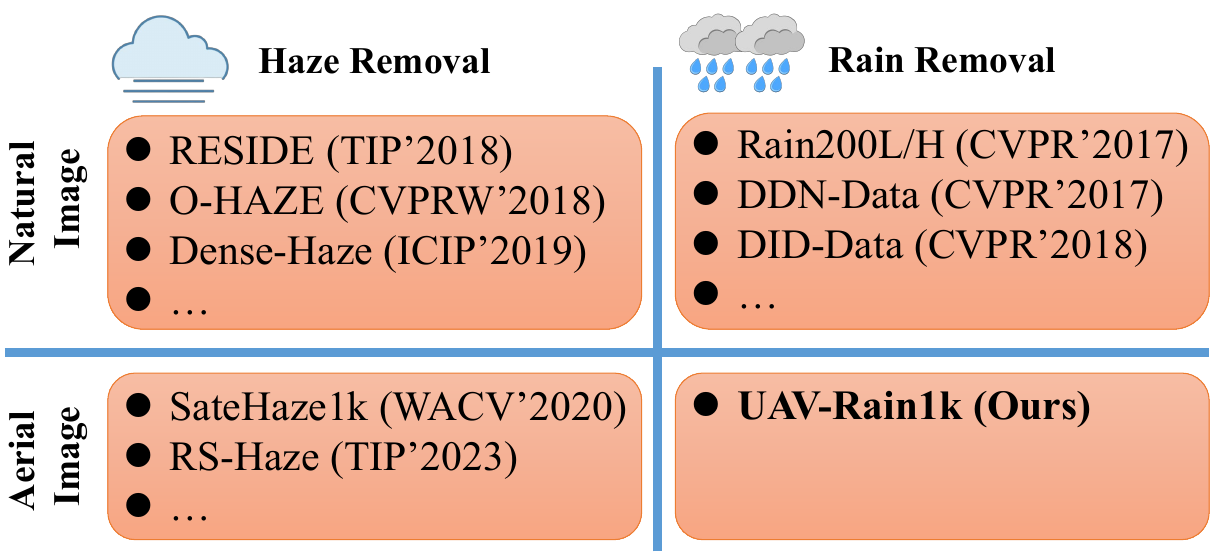}
\caption{Classification diagram of publicly available datasets for haze removal and rain removal. Our proposed UAV-Rain1k fills the research gap.}
\label{fig1}
\end{figure}

Although the above mentioned datasets \cite{qian2018attentive, quan2021removing} have propelled research in raindrop removal, they are all tailored for autonomous driving scenarios. When revisiting research on UAV image processing under adverse weather conditions, most existing studies focus on haze removal, with little attention given to raindrop removal tasks. However, the reality is that raindrop interference on UAV aerial images is also a common scenario in rainy conditions. Compared to autonomous driving scenarios, raindrop effects on UAV aerial imagery pose greater challenges. On one hand, UAVs operate in dynamic environments, which can lead to unpredictable raindrop patterns on lenses or sensors. On the other hand, UAVs often capture images from varying distances and angles, resulting in inconsistent sizes and shapes of raindrops within the images.
\begin{figure*}[!t]
\centering
\includegraphics[width=1.0\textwidth]{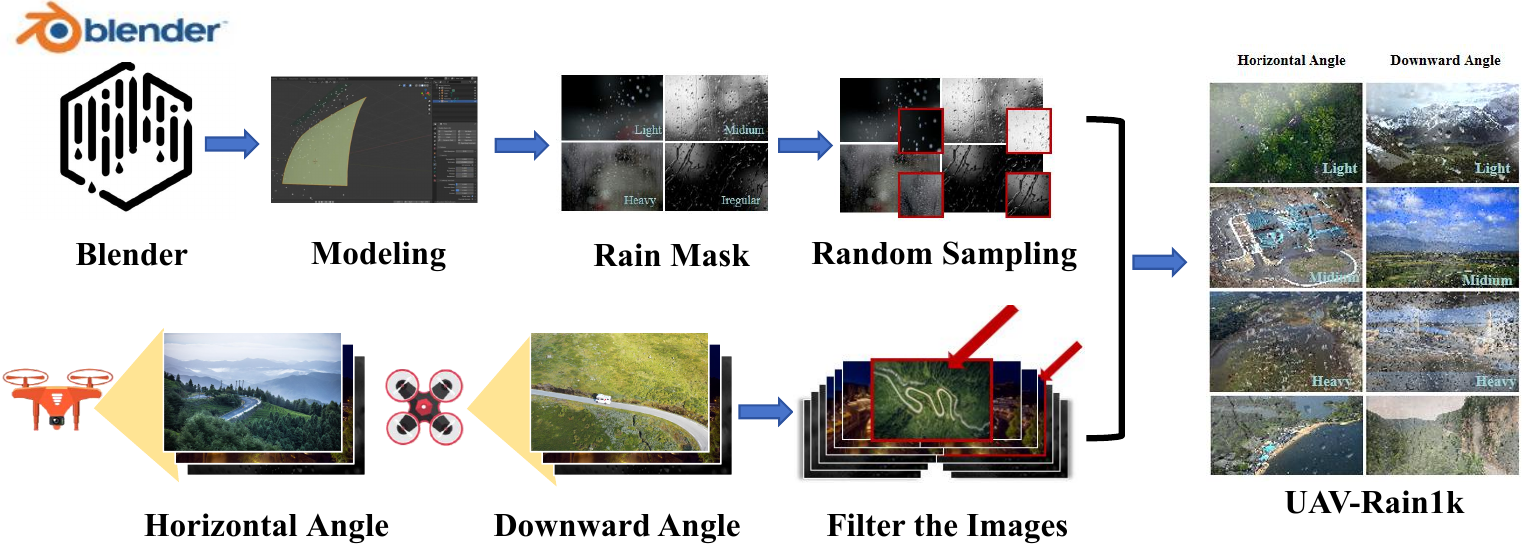}
\caption{Illustration of the dataset generation pipeline.}
\label{fig2}
\end{figure*}

To fill the gap in this research, we present a new benchmark for raindrop removal from UAV aerial imagery, called UAV-Rain1k. To the best of our knowledge, this is the first dataset specifically curated for raindrop removal tasks in the field of aerial image processing (see Figure \ref{fig1}). In addition, we further conduct a comprehensive evaluation of several state-of-the-art image deraining methods, which are quantitatively and qualitatively evaluated on our new dataset. Our evaluation and analysis demonstrate the performance and limitations of existing approaches, providing promising insights. The proposed UAV-Rain1k dataset is publicly available for research purposes, and we intend to periodically update our benchmarking results for noticeable new algorithms.

The rest of this paper is structured as follows. In Section~\ref{sec2}, we provide a detailed description of the pipeline used to construct the dataset. We analyze the performance of existing algorithms on our benchmark in Section \ref{sec3}. Finally, the concluding remarks will be given in Section \ref{sec4}.

\section{Dataset Construction}
\label{sec2}
In this section, we describe the process of dataset construction. Since obtaining paired real-world raindrop and rain-free aerial images for the same scene and field of view is not feasible, we synthesize a more diverse dataset of UAV aerial images with raindrops. Due to the diverse scenes, complex objects, and vast spatial coverage in aerial images, the previous methods \cite{li2019single} for synthesizing rainy images are not suitable for aerial imagery. For instance, in the previously synthesized data \cite{qian2018attentive,quan2021removing}, raindrops are consistently simulated as being captured at a horizontal angle over land. However, in reality, the distribution and size of raindrop adhesions vary with different aerial angles. Therefore, we construct a dataset that takes into account the realistic distribution of raindrops in aerial scenes. Figure \ref{fig2} shows the dataset generation pipeline. The details will be discussed below.

\subsection{Raindrop Generation}
\label{sec:2.1}
As a complex atmospheric process, rainfall can lead to various forms of visibility reduction due to environmental factors such as raindrop size, rain density, and wind speed. The fidelity and intricacy of rain play a crucial role in the synthesis of raindrops \cite{quan2019deep}. To achieve higher quality and more realistic synthetic aerial images with raindrops, we employ an open-source 3D graphics engine (Blender) to simulate and generate images of raindrops within real aerial scenes for training purposes. This 3D graphics engine is capable of rendering raindrops using a physical motion model, allowing us to set depth information and color values separately in the RGB channels \cite{huang2021single}, which is facilitated by the Blender plugin known as the Rain Generator. 

Inspired by Garg \cite{garg2006photorealistic}, we model the generation of raindrop layers as a motion blur process. The instantaneous shape of a water droplet at time $t$ is represented by the function $r[t,\vartheta ,\phi ]$, where $r$ is the distance from a point on the droplet's surface to its center, $\theta$ is defined as the angle opposite to the direction of the droplet's fall at that point, and $\phi$ is defined as the angle between that point and the projection of the line of sight onto any plane perpendicular to the direction of fall. As raindrops fall, the influence of aerodynamics and surface tension acting on the raindrop causes rapid shape distortion over time. Mathematically, this motion process can be represented based on the oscillatory characteristics of the falling raindrop:
\begin{equation}
r[t,\vartheta ,\phi ]={r}_{0}(1+\displaystyle\sum_{m}\cos (m\phi){p}_{m}(\theta )),
\end{equation}
where ${r}_{0}$ is the undeformed radius, and the factor $\cos (m\phi)$ depends on the droplet size ${r}_{0}$. The function ${p}_{m}(\theta )$ describes the time dependent variation of modal shape and amplitude with respect to $\theta$. The droplet's shape at any given moment is determined by the cumulative impact of falling time on both modal shape and amplitude.

\subsection{Background Collection}
\label{sec:2.2}
The quality of the clear background image, also known as the ground-truth image, is equally crucial when constructing a paired dataset \cite{guo2023sky,chen2023towards,li2022toward}. These images should have higher resolution and rich object details. Unfortunately, existing datasets have placed significant emphasis on rain simulation but have overlooked the need for high-quality background images. Regrettably, the realistic rainy weather images used in previous synthetic datasets were consistently captured from a flat angle on the ground. In reality, there is a significant disparity between the lighting conditions in the air and those on the ground, especially in aerial imagery. Aerial images predominantly feature ground structures as the primary compositional elements in the background, which aligns with human visual perception of rainy scenes. Obtaining rain-free background images that accurately represent the human perception of real rainy conditions is a challenging task, particularly when searching for such images on the Internet. These images also suffer from various issues, including compression artifacts, watermarks, low resolution, defocused blurriness, absence of objects, and etc. These challenges can present difficulties for advanced vision-based remote sensing applications.
\begin{figure}[!t]
\centering
\includegraphics[width=1.0\columnwidth]{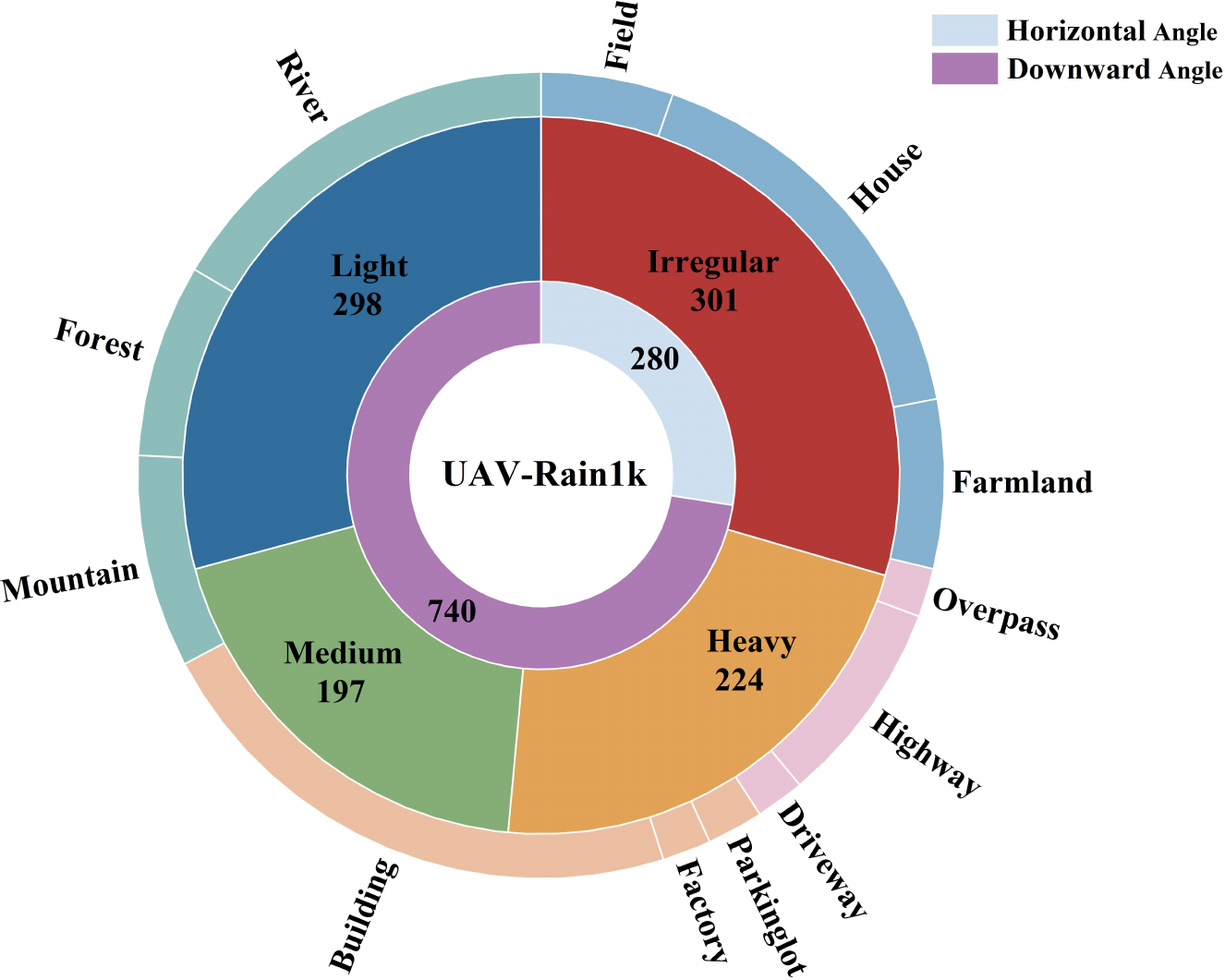}
\caption{Distribution of rain and scene of the proposed benchmark.}
\label{fig3}
\end{figure}
\begin{figure}[!t]
\centering
\includegraphics[width=1.0\columnwidth]{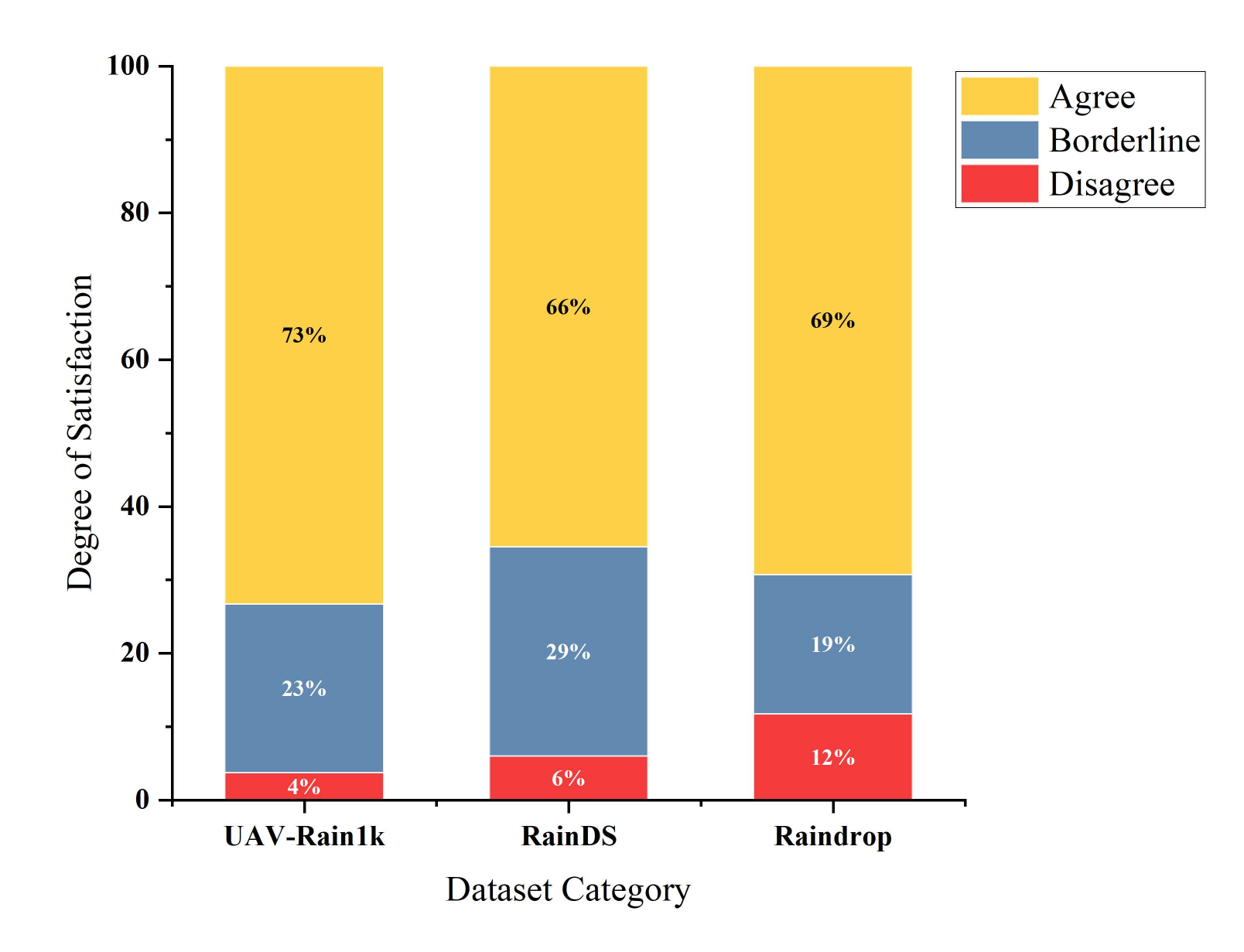}
\caption{User study results. The ratings given by all participants on different raindrop datasets.}
\label{fig4}
\end{figure}

Our collection of background images encompasses a wide range of scenes, including urban, natural, and rural environments. Furthermore, we deliberately sought variations in aerial shooting angles during the photography process to capture different scene effects in real scenarios. Consequently, the complexity of these scenes and backgrounds presents both challenges and value to our subset of real data.
\begin{figure*}[!t]
\centering
\includegraphics[width=1.0\textwidth]{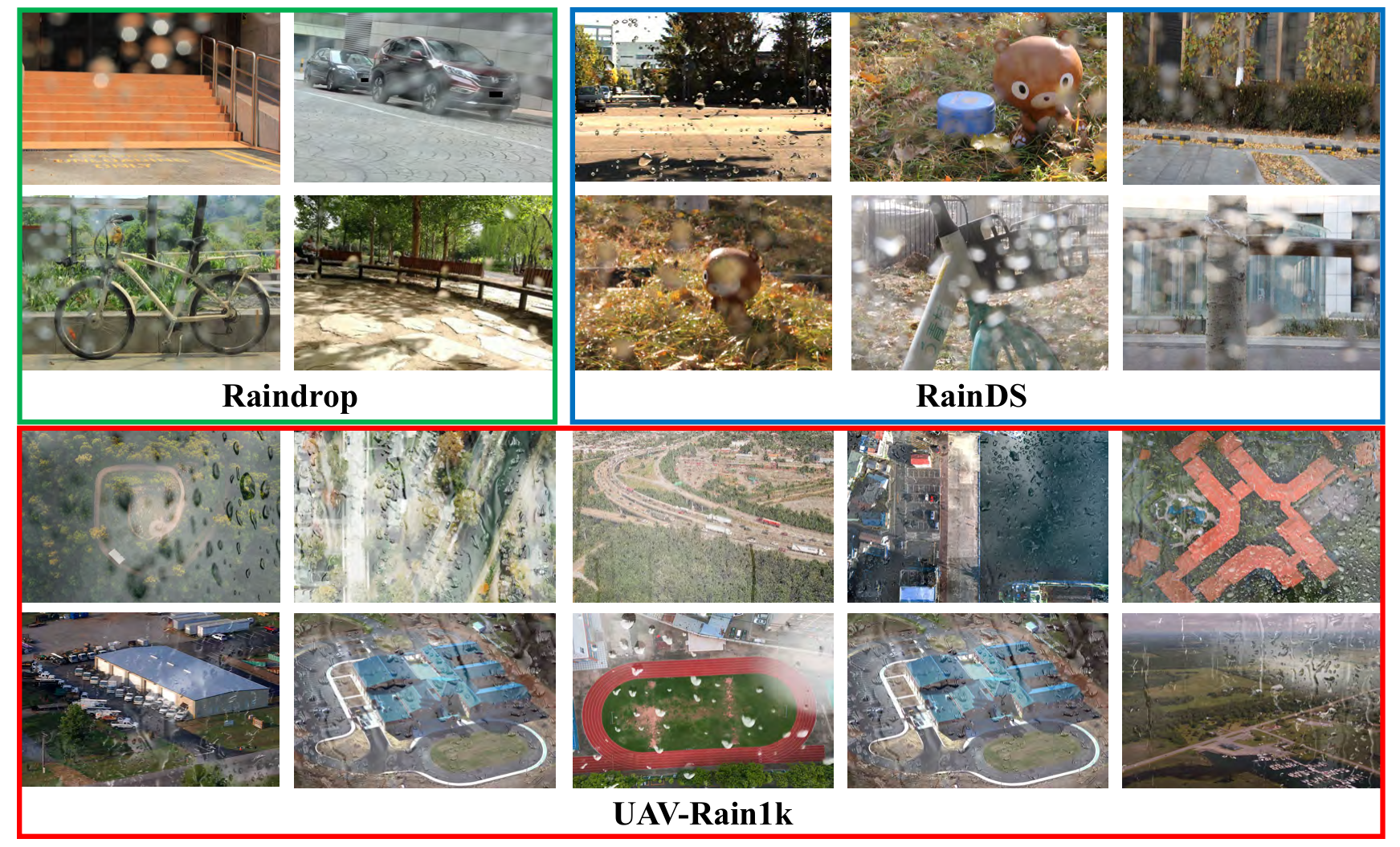}
\caption{Example images from previous representative datasets (Raindrop \cite{qian2018attentive} and RainDS \cite{quan2021removing}) and our proposed more challenging UAV-Rain1k.}
\label{fig5}
\end{figure*}
\begin{table*}[t]
\centering
\caption{Quantitative comparisons on the UAV-Rain1k benchmark dataset. ``\#FLOPs'' and ``\#Params'' represent FLOPs (in G) and the number of trainable parameters (in M), respectively.}
\resizebox{0.95\textwidth}{!}{
\begin{tabular}{cccccccc}
\hline
Methods   & Input  & DSC \cite{luo2015removing}   & RCDNet \cite{wang2020model} & SPDNet \cite{yi2021structure} & Restormer \cite{zamir2022restormer}  & IDT \cite{xiao2022image}        & DRSformer \cite{chen2023learning}  \\ \hline
Category  & -      & Prior  & CNN    & CNN    & Transformer & Transformer & Transformer \\
PSNR      & 16.80  & 16.68  & 22.48  &22.54   &24.78        & 22.47       &   24.93     \\
SSIM      & 0.7196 & 0.7142 & 0.8753 &0.8594  & 0.9054      & 0.8957      & 0.9155      \\
\#Params  & -      & -      & 3.17   & 3.04   & 26.12       & 16.41       & 33.65       \\
\#FLOPs  & -      & -      & 21.2   & 89.3   & 174.7       & 61.9        & 242.9       \\ \hline
\end{tabular}
}
\label{table1}	
\end{table*}

To generate synthetic data, we employ the SMRC-ct13 model unmanned aerial vehicle for on-site shooting. To ensure that the subsequently synthesized rainy images are more realistic and cohesive, we deliberately select overcast and rainy weather conditions for capturing background materials. Additionally, during the shooting process, we diversify our approach by flying the drone in different directions to cover scenes from various angles. Due to constraints in selecting actual shooting locations for aerial images, we complement our dataset by acquiring copyright-free and non-commercial aerial images and videos through Google searches to synthesize rainy weather images. In total, we capture and collect 327 aerial video sequences, comprising approximately 25,000 frames.

To maintain the richness and diversity of our dataset, we carefully select frames from each video sequence that met our criteria, filtering out low-quality images containing issues such as low resolution, website watermarks, compression artifacts, and blurriness. We also exclude background images with excessive lighting or high brightness to ensure appropriate background composition. In summary, we capture and collect footage from over 12 typical scenes, including parking lots, streets, alleys, playgrounds, courtyards, and forests. The solar storm diagram as shown in Figure~\ref{fig3}.

\subsection{Image Composition}
\label{sec:2.3}
In \cite{qian2018attentive}, a method is employed where a piece of glass with attached water droplets is inserted in front of the drone camera lens to simulate the presence of raindrops. However, during the process of capturing aerial rain images, due to the complexity and specificity of capturing scenes with moving perspectives, we are unable to obtain perfectly consistent rainy and rain-free images through aerial images. This method also provides some insights for verifying the authenticity of synthetic datasets. We collect real aerial images captured during rainy weather and gather drone footage of raining scenes at different angles and motion states as a reference baseline for synthetic images. This ensures that the collected images exhibit variations in raindrop density, size, and positions, allowing us to filter out low-quality and misleading synthetic data.

Our goal is to achieve visual realism and consistency in synthetic rainy images, with the aim of minimizing the domain gap between synthetic and real images. To achieve this, we do not rely on simple copy-and-paste methods. Instead, we employ a random sampling approach for raindrop masks to ensure that each synthesized rainy image is unique. To ensure the synthesized images exhibit visual realism and reduce the domain gap between synthesized and real images, inspired by Qian et al. \cite{qian2018attentive}, we model the blurred or occluded effects $D$ of raindrops in dispersed small-sized locally coherent regions, combined with a clean background image $B$, to form the degraded raindrop image ${R}_{d}$. Mathematically, the synthesis process can be expressed as follows:
\begin{equation}
{R}_{d}=(1-M)\bigodot B+D,
\end{equation}
where $D$ represents the occlusion or blurring effects caused by raindrops, and $B$ is a binary mask. If $M(x)=1$, then the pixel $x$ in the mask belongs to the raindrop region, otherwise, it is part of the background image.
As a result, we seamlessly integrate the raindrop mask with the background image. Consequently, compared to other dataset generation methods, our proposed dataset tends to be more complex and diverse, enhancing its ability to generalize across various scenarios.

\subsection{Benchmark Statistics}
\label{sec:2.4}
As a result, we propose a new benchmark for raindrop removal from UAV aerial imagery, called UAV-Rain1k. In total, the training and testing set of the UAV-Rain1k contains 800 and 220 synthetic images, respectively. The average resolution of all images is $1500 \times 1000$. Our dataset comprises four rain density labels (e.g., light, moderate, heavy, and irregular). In Figure \ref{fig3}, the inner ring displays labels for each rain density, corresponding to a specific quantity of synthesized images. Furthermore, we categorize the background images according to aerial shooting angles and aerial scenes. Consequently, our dataset demonstrates greater diversity and a larger scale in terms of both rain conditions and ground truth (GT) images. Our emphasis goes beyond just the sequence number and total frames of real rain. We also prioritize the diversity of real rain patterns and the authenticity of background images. This focus has a significant impact on the model's ability to generalize to real rain scenarios. 

Based on the aforementioned work, we have summarized three innovative aspects of the UAV-Rain1k dataset:
\begin{compactitem}
  \item 
  A more comprehensive and reliable background collection effort. Encompassing a diverse range of scenes, while simultaneously considering the complexity and uniqueness of capturing scenes from dynamic perspectives.
  \item 
 More authentic raindrop shapes. We model the generation of the raindrop layer as a motion blur process and employed an open source 3D rendering engine for the depiction of raindrops, achieving higher quality and more realistic raindrop masks.
  \item
  A more natural synthesis effect. Integrating authentic and diverse raindrop masks seamlessly with background images covering various scenes to achieve a more realistic visual effect. Figure \ref{fig5} provides some visual examples.
\end{compactitem}

\begin{figure*}[!t]
\centering
\includegraphics[width=1.0\textwidth]{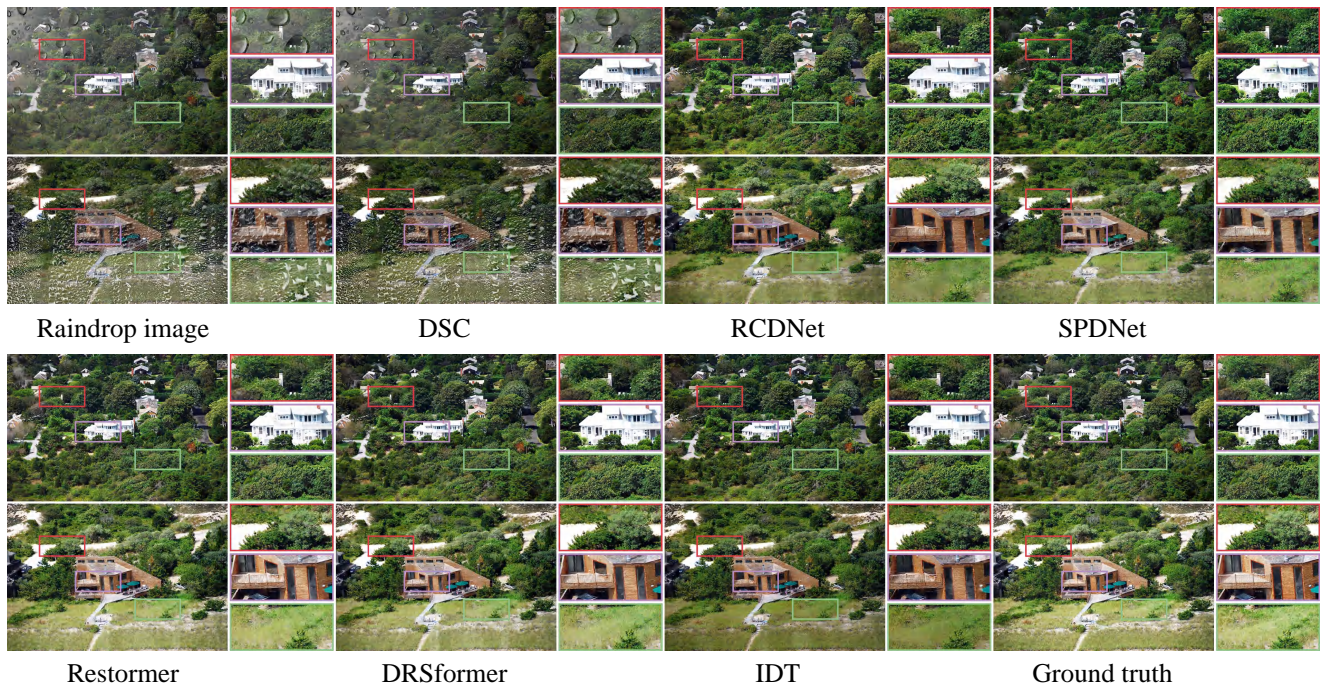}
\caption{Visual quality comparison on the UAV-Rain1k dataset (horizontal angle). Zooming in the figures offers a better view at the deraining capability.}
\label{fig6}
\end{figure*}
\begin{figure*}[!t]
\centering
\includegraphics[width=1.0\textwidth]{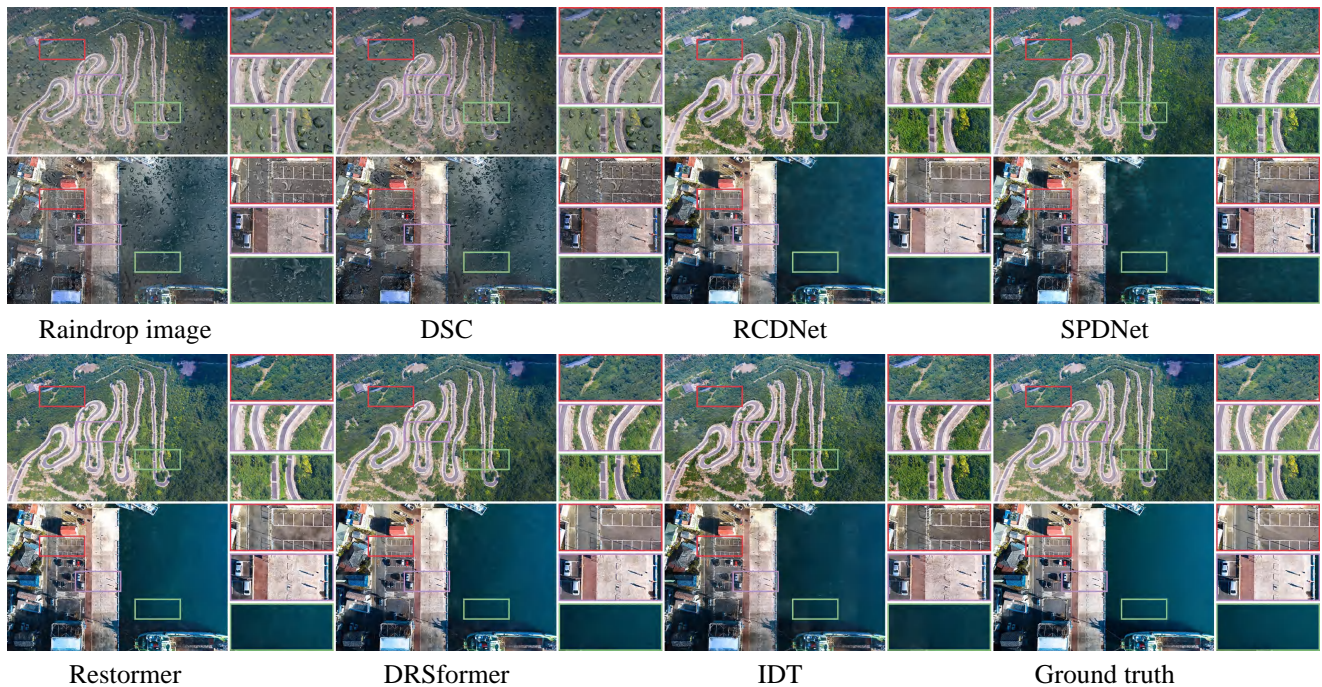}
\caption{Visual quality comparison on the UAV-Rain1k dataset (downward angle). Zooming in the figures offers a better view at the deraining capability.}
\label{fig7}
\end{figure*}
\begin{figure*}[!t]
\centering
\includegraphics[width=1.0\textwidth]{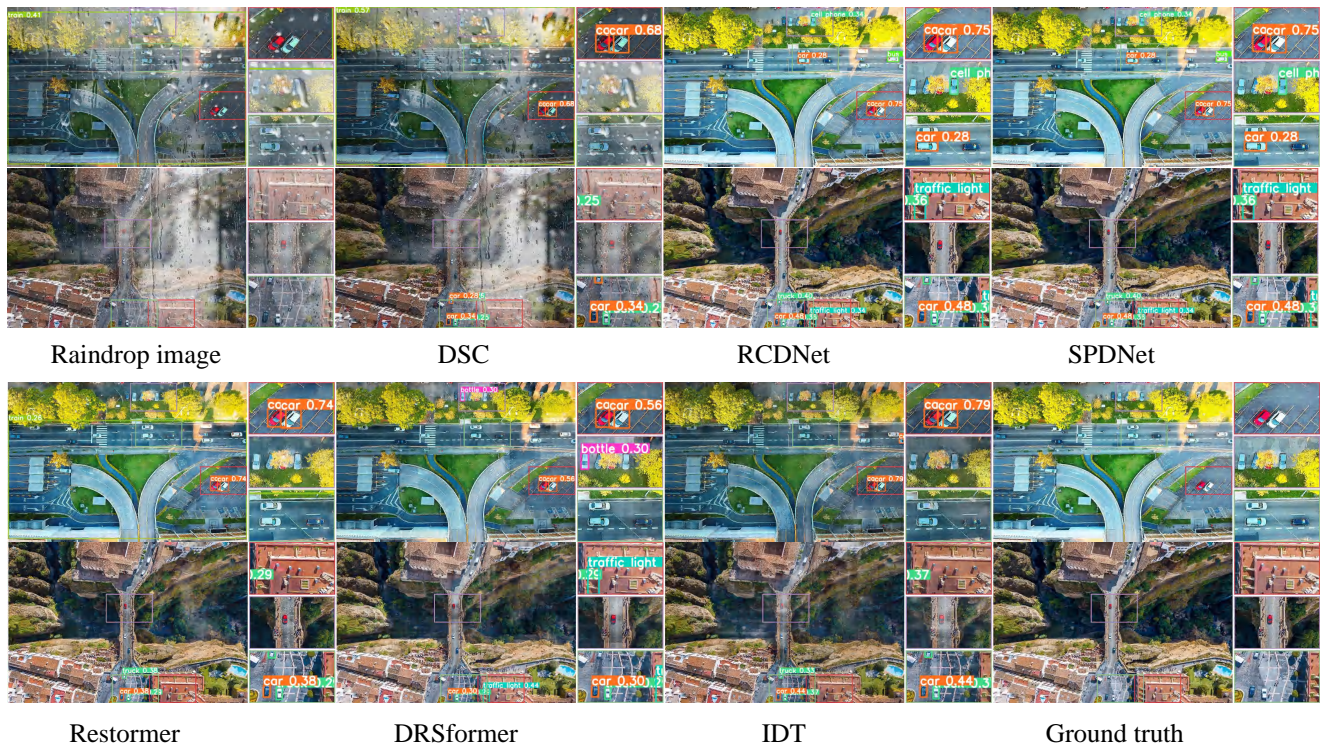}
\caption{Visualization of the object detection results after applying different deraining algorithms.}
\label{fig8}
\end{figure*}

\subsection{Subjective Assessment}
\label{sec:2.5}
We conduct an online user study to evaluate the quality of the synthesis rainy images. Here, we prepare for 90 rainy images, randomly chosen from 3 datasets (Raindrop \cite{qian2018attentive}, RainDS \cite{quan2021removing} and UAV-Rain1k) with 30 samples from each dataset. We recruit 40 participants, consisting of 20 males and 20 females. Each participant is shown a random set of 150 rainy images. Then, employing a 3-point Likert scale (i.e., agree, neutral, disagree), all participants assess the perceived realism of each image. In the end, each category receives 600 ratings, which are used to assess the quality of the synthetic dataset. Subjective scores are shown in Figure \ref{fig4}, where our UAV-Rain1k consistently outperforms the other two benchmarks. This also indicates that our synthesized rain is assessed as more realistic compared to previous datasets. Note that we do not standardize the scores. Therefore, what matters here is the score ranking rather than the absolute score values. Despite the relatively small number of evaluators, we observe good consensus and small inter-rater differences in the same paired comparison results, making the scores reliable.

\section{Algorithm Benchmarking}
\label{sec3}
In this section, based on the proposed new benchmark, we evaluate 6 representative image deraining algorithms: DSC \cite{luo2015removing}, RCDNet \cite{wang2020model}, SPDNet \cite{yi2021structure}, Restormer \cite{zamir2022restormer}, IDT \cite{xiao2022image}, and DRSformer \cite{chen2023learning}. To ensure fair comparisons, we utilize the officially released codes of these approaches. Each method undergoes retraining on the UAV-Rain1k benchmark, conducted on servers equipped with NVIDIA GeForce RTX 3090 GPUs.

\subsection{Quantitative Evaluation}
\label{sec:3.1}
Table \ref{table1} presents the quantitative performance evaluation results of various algorithms on the UAV-Rain1k dataset. Here, we utilize two widely referenced image quality assessment metrics, PSNR (Peak Signal-to-Noise Ratio) \cite{huynh2008scope} and SSIM (Structural Similarity Index) \cite{wang2004image}, to measure the restoration quality of different methods. Additionally, we assess the model efficiency of each approach by considering parameters and FLOPs (Floating Point Operations per Second).

It is evident from the results that DRSformer \cite{chen2023learning} and Restormer \cite{zamir2022restormer} achieve the top two quantitative outcomes in rain removal performance. However, it is noteworthy that their model complexity is comparatively higher than traditional CNN methods \cite{wang2020model,yi2021structure}. To comprehensively evaluate the performance and efficiency of different algorithms, future research endeavors could delve deeper into strategies for maintaining high performance while reducing model complexity. This would address the requirements of speed and resource consumption in real-world remote sensing applications \cite{chen2022unpaired}.


\subsection{Qualitative Evaluation}
Figure \ref{fig6} and Figure \ref{fig7} illustrates the visual comparison results of different baseline algorithms on our proposed benchmark (In Figure \ref{fig6}, the effect corresponds to a composite image taken from a horizontal angle, while Figure~\ref{fig7} represents a composite image captured from an downward angle). In comparison to CNN-based methods (RCDNet \cite{wang2020model} and SPDNet \cite{yi2021structure}), Transformer-based approaches exhibit superior capability in removing undesired raindrop and recovering clearer visual outcomes, it achieves remarkable visual effects from any aerial perspective, eliminating raindrops and obscuring effects without rain residues and pseudo artifacts. owing to the evident advantage of Transformers in modeling non-local information. Additionally, we observe grid artifacts in the restoration results of IDT \cite{xiao2022image}, indicating their limited capacity in handling high-resolution UAV imagery. Furthermore, nearly all methods exhibit poor performance in local detail regions, as they tend to remove image details while removing raindrop, suggesting that there are still rooms for improvement.

\subsection{Application-Based Evaluation}
To investigate whether the raindrop removal process benefits downstream vision-based remote sensing applications, we apply the popular object detection pre-trained model (YOLOv8 \cite{wang2023uav}) to evaluate the deraining results. Figure~\ref{fig8} shows the visualization of the object detection results after applying different deraining methods. As one can see, the detection precision of the deraining results by all approaches demonstrates varying degrees of improvement compared to the input raindrop images, especially in the recognition of vehicles on the road. While removing the raindrop regions, some existing methods may also remove the semantic information, disturbing human understanding and degrading high-level vision algorithms. Future works could integrate low-level and high-level tasks to ensure the former contributes to the latter.

\section{Conclusion}
\label{sec4}
In this paper, we have proposed a new benchmark dataset for raindrop removal from UAV aerial imagery. We have provided a detailed overview of the synthesis process for the UAV-Rain1k dataset, covering background collection, raindrop modeling, image synthesis, results demonstration, and subjective evaluation. Based on the proposed benchmark, we further experimentally validated representative image deraining algorithms in three aspects: quantitative evaluation, qualitative evaluation, and application-based evaluation. Experimental results shed light on the limitations of existing methods and point towards promising future directions. This paper not only pioneers a new research direction in UAV image processing but also advocates for the community to propose more effective algorithms. 

{
    \small
    \bibliographystyle{ieeenat_fullname}
    \bibliography{main}
}

\end{document}